\documentclass[letterpaper, 10 pt, conference]{ieeeconf}  

\IEEEoverridecommandlockouts                              

\usepackage{colortbl}
\usepackage{diagbox}
\usepackage{epsfig}   
\usepackage{amsmath}  
\usepackage{amssymb}  
\usepackage{color}
\usepackage{dblfloatfix}
\usepackage[linesnumbered,ruled]{algorithm2e}
\usepackage[normalem]{ulem} 
\makeatletter
\let\NAT@parse\undefined
\makeatother
\usepackage{hyperref}
\hypersetup{pdfstartview=FitH,
            colorlinks=true,
            linkcolor=red,
            anchorcolor=blue,
            citecolor=green
            }
\usepackage{cite}
\usepackage{booktabs}
\usepackage{multirow}
\usepackage[caption=false,font=footnotesize]{subfig}
\usepackage[dvipsnames]{xcolor}
\definecolor{table_c}{RGB}{248,243,249}
\usepackage{soul}
\usepackage{balance}
\newcounter{RNum}
\usepackage{bm}
\renewcommand{\theRNum}{\arabic{RNum}}
\newcommand{\Remark}{\noindent\textit{\textbf{Remark}~\refstepcounter{RNum}\textbf{\theRNum}: }}
\newcommand{\NoOne}[1]{\textcolor{red}{#1}}
\newcommand{\NoTwo}[1]{\textcolor{green}{#1}}
\newcommand{\NoThree}[1]{\textcolor{blue}{#1}}
\soulregister\NoOne7
\soulregister\NoTwo7
\soulregister\NoThree7
\soulregister\Remark7

\title{\LARGE \bf
Progressive Representation Learning for Real-Time UAV Tracking
}

\author{Changhong Fu$^{1*}$, Xiang Lei$^{2}$, Haobo Zuo$^{3}$, Liangliang Yao$^{1}$, Guangze Zheng$^{3}$,  and Jia Pan$^{3}$ 
\thanks{*Corresponding author} 
\thanks{$^{1}$C. Fu and L. Yao are with the School of Mechanical Engineering, Tongji University, Shanghai 201804, China. \itshape{Email: changhongfu@tongji.edu.cn}} %
\thanks{$^{2}$X. Lei is with the School of Software Engineering, Tongji University, Shanghai 201804, China.} %
\thanks{$^{3}$H. Zuo, G. Zheng, and J. Pan are with the Department of Computer Science, the University of Hong Kong, Hong Kong, China.} %
}

\begin{document}

\maketitle
\thispagestyle{empty}
\pagestyle{empty}

\begin{abstract}
Visual object tracking has significantly promoted autonomous applications for unmanned aerial vehicles (UAVs).
However, learning robust object representations for UAV tracking is especially challenging in complex dynamic environments, when confronted with aspect ratio change and occlusion.
These challenges severely alter the original information of the object.
To handle the above issues, this work proposes a novel progressive representation learning framework for UAV tracking, \textit{i.e.}, PRL-Track.
Specifically, PRL-Track is divided into coarse representation learning and fine representation learning. 
For coarse representation learning, two innovative regulators, which rely on appearance and semantic information, are designed to mitigate appearance interference and capture semantic information.  
Furthermore, for fine representation learning, a new hierarchical modeling generator is developed to intertwine coarse object representations.
Exhaustive experiments demonstrate that the proposed PRL-Track delivers exceptional performance on three authoritative UAV tracking benchmarks.
Real-world tests indicate that the proposed PRL-Track realizes superior tracking performance with 42.6 frames per second on the typical UAV platform equipped with an edge smart camera.
The code, model, and demo videos are available at \url{https://github.com/vision4robotics/PRL-Track}.
\end{abstract}

\section{Introduction}
Robust visual object tracking is fundamental for intelligent unmanned aerial vehicle (UAV) applications,~\textit{e.g.}, task planning~\cite{10341486}, biodiversity protection~\cite{10341725}, and target localization~\cite{mcarthur2020pose}. 
During the above extensive applications, UAV trackers aim to predict the location of the object in subsequent frames, starting from the initial position in the first frame.
Driven by large-scale datasets with manual annotations, Siamese trackers~\cite{cao2021siamapn++, fu2023siamese, wang2019unsupervised, fu2024SAMDA} have shown promising performance by adopting convolutional neural networks (CNNs) to learn object representations.
However, when encountered with complex dynamic environments, \textit{e.g.}, aspect ratio change and occlusion, these trackers struggle to obtain robust object representations due to limited representation capabilities of lightweight CNNs like AlexNet~\cite{krizhevsky2012imagenet}.
Although trackers with deeper backbones,~\textit{e.g.}, ResNet~\cite{he2016deep}, can better learn object representations, they fail to meet the real-time requirement constrained by limited computational resources on UAVs.
\textbf{\textit{Hence, robust object representations for UAV tracking are far from sufficient in complex dynamic environments.}}

\begin{figure}[!t]	
    \centering
    \vspace{6pt}
    \colorbox{table_c}{\includegraphics[width=1\linewidth]{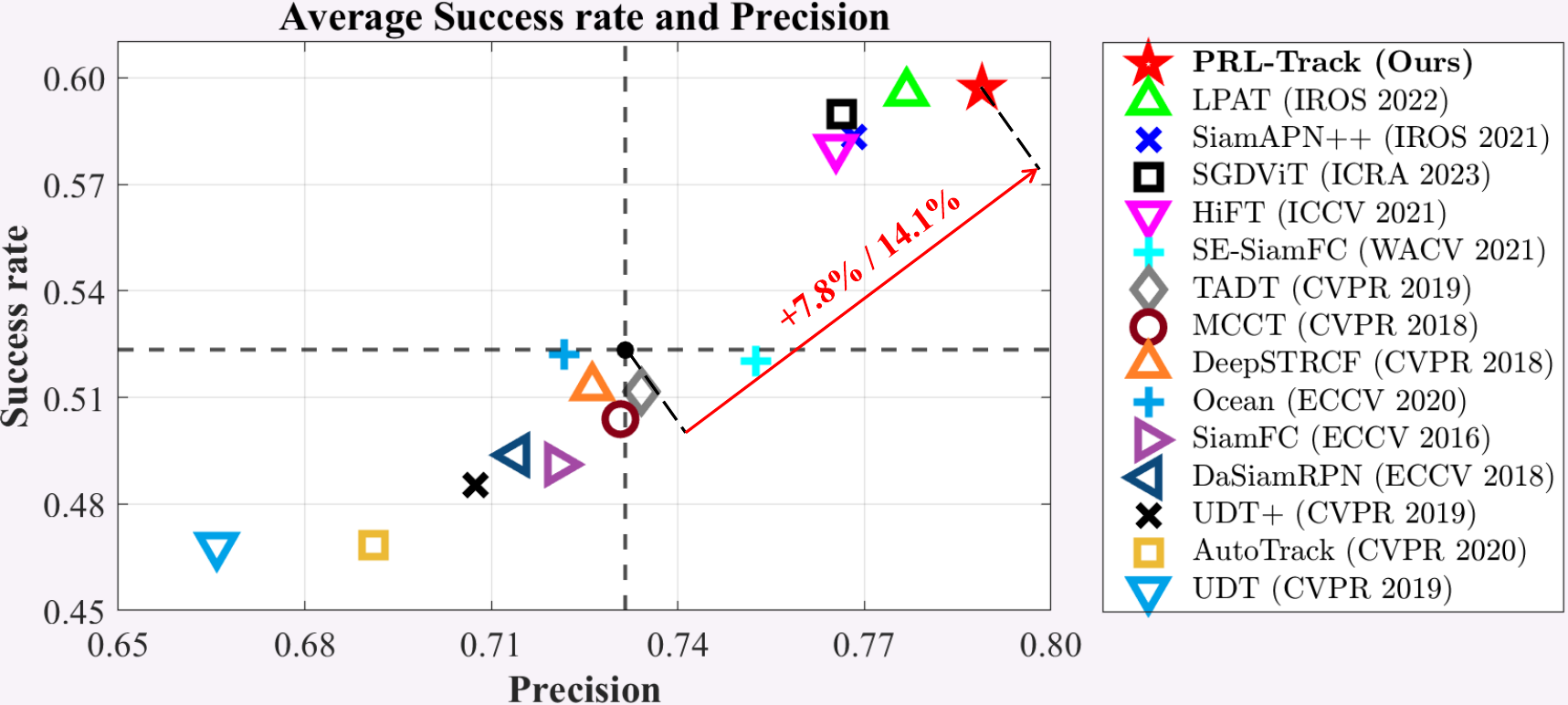}}
    \caption{Overall comparison of the proposed PRL-Track with other 14 state-of-the-art (SOTA) trackers on the combination of UAV tracking benchmarks. PRL-Track achieves more robust performance than other 14 SOTA trackers. Specifically, PRL-Track surpasses the average precision and success rate of the 14 trackers (\textbf{black dot}) by \textbf{7.8\%} and \textbf{14.1\%}, respectively.}
    \label{fig:sg1}
    \vspace{-15pt}
\end{figure}

One promising approach is to explore multi-scale features oriented to UAV tracking tasks~\cite{cao2021siamapn++}. 
Specifically, convolutional operations are adopted to aggregate multi-scale features from different layers, which contribute to alleviating feature degradation due to occlusion during UAV tracking.
However, with limited receptive fields of convolutional kernels, CNNs lack the modeling ability of long-range dependencies~\cite{9857212}.
Consequently, it is challenging to capture global context information between multi-scale features.
Recently, Vision Transformer (ViT)~\cite{dosovitskiy2020image} has exhibited tremendous potential in modeling long-range dependencies by virtue of attention mechanisms.
The introduction of ViT into the Siamese trackers addresses the shortcomings of traditional CNN-based trackers in learning global information. 
Moreover, the intrinsic global modeling capability of ViT proves to be advantageous in tackling appearance variations, \textit{e.g.}, aspect ratio change~\cite{cao2021hift}.
Nonetheless, compared with CNN, ViT tends to ignore local spatial information, which decreases the discriminability of image objects~\cite {10040235}.
Besides, the quadratic computational complexity and memory cost of the attention mechanisms are obstacles to its wide deployment on embedded processors in UAVs, which have limited computing resources.
\textbf{\textit{Therefore, how to extract more reliable information and then generate robust object representations for UAV tracking is worth exploring carefully.}}

To fully exploit the global context information and local spatial information, integrating CNNs and ViT represents a promising complementary coupling.
Given the strength of CNNs in fast convergence and filtering redundant information~\cite{8546079, 8850096}, they are well-suited for extracting object local information from images to form coarse object representations.
Subsequently, ViT utilizes coarse object representations to refine and enhance the understanding of global context information, thereby generating robust fine object representations.
However, considering the distinctions in feature space between the plain CNNs and ViT, directly concatenating them leads to performance degradation~\cite{10143709, raghu2021vision}.
\textbf{\textit{Therefore, how to effectively integrate CNNs and ViT for real-time UAV tracking is a problem worth exploring.}}

This work proposes a novel progressive representation learning framework, namely PRL-Track, which consists of CNN-based coarse representation learning and ViT-based fine representation learning.
Leveraging the complementary strengths of the CNNs and ViT, PRL-Track can learn robust fine object representations, achieving satisfactory performance when encountering challenges such as occlusion and aspect ratio change during UAV tracking.
Fig.~\ref{fig:sg1} highlights the impressive performance of PRL-Track in UAV tracking, outperforming other 14 state-of-the-art (SOTA) trackers in terms of average precision and success rate.
The main contributions of this work are as follows:
\begin{itemize}
\item A novel progressive representation learning framework dubbed PRL-Track, is proposed to learn robust fine object representations for UAV tracking via a coarse-to-fine perspective, thus improving tracking performance.
\item An innovative appearance-aware regulator is developed to mitigate appearance interference and extract useful information from shallow features for coarse representation learning. Besides, a convenient semantic-aware regulator is designed to capture semantic information and promote the concentration of deep features.
\item A new hierarchical modeling generator is proposed to augment the comprehension of contextual information by fusing coarse object representations for fine representation learning, further generating robust fine object representations for UAV tracking.
\item Comprehensive evaluations confirm that PRL-Track achieves SOTA performance, validating the power of the proposed framework. Real-world tests conducted on the typical UAV platform demonstrate the superior efficiency and robustness of PRL-Track in practical scenarios.
\end{itemize}

\section{Related Work}
\subsection{UAV Tracking}
Siamese trackers~\cite{cao2021siamapn++, bertinetto2016fully, guo2017learning} have gained popularity and promoted the development of UAV applications owing to their remarkable tracking performance.
These trackers utilize a CNN-based backbone to extract features of both the template patch and search patch, followed by a correlation-based network to calculate the similarity between them.
Compared with correlation filter-based trackers~\cite{li2020autotrack, wang2018multi}, fully CNN-based trackers further exploit the local spatial information, thus improving tracking performance.
As a pioneer, SiamFC~\cite{bertinetto2016fully} introduces the Siamese framework into object tracking for similarity matching.
Inspired by the region proposal network, SiamRPN~\cite{li2018high} combines a Siamese network with regression and classification branches, achieving efficient classification and accurate prediction.
However, trackers with fully CNN-based architecture lack effective long-range dependency modeling, which means they often struggle to capture global context information.
Thus, it is difficult to ensure reliable tracking in complex dynamic environments.
To address this issue, ViT~\cite{dosovitskiy2020image}  has been introduced into object tracking, owing to its high representational capacity for global context information.
ViT integrates global contextual information by decomposing the image into fixed-size blocks and processing them with Transformer architecture.
TransT~\cite{chen2021transformer} proposes a ViT-based feature fusion model for object tracking, achieving promising performance.
HiFT~\cite{cao2021hift} introduces a ViT structure optimized for efficient multi-feature fusion, thereby augmenting tracking robustness.
SGDViT~\cite{10161487} designs a saliency-guided dynamic ViT to capture similarity and incorporate information.
However, the attention mechanism in ViT often ignores local feature details and object spatial structures~\cite{10040235}.
Therefore, a promising approach to overcome these limitations is the integration of CNNs and ViT, leveraging the strengths of both architectures in the context of UAV tracking.
CNNs can capture local spatial information, which contributes to maintaining accuracy in the rapid environments in which UAVs operate.
By integrating this with ViT's ability to model global context, the framework can better understand broader scene dynamics, enabling more stable tracking across wide fields of view.

\subsection{Representation Learning}
Representation learning aims to acquire object representations that facilitate the utilization of reliable information when constructing classifiers or predictors~\cite{6472238}.
Deep neural networks (DNNs) are commonly employed to extract object representations in visual tasks~\cite{krizhevsky2012imagenet, zou2023object}.
Compared with conventional hand-crafted representations, DNNs tend to learn more comprehensive representations~\cite{wang2020deep}.
Previous works on representation learning have yielded notable frameworks and methodologies.
UniFormer~\cite{10143709} designs a concise unified framework and integrates the strength of CNNs and the ViT, realizing efficient spatiotemporal representation learning.
EsViT~\cite{li2021esvit} formulates an efficient self-supervised ViT for representation learning, achieving superior transfer performance in downstream tasks.
MARLIN~\cite{cai2023marlin} employs a facial video masked autoencoder to learn generic and robust facial representations.
HRNet~\cite{wang2020deep} proposes to uphold high-resolution object representations throughout the entire workflow, thereby ensuring the reliability of object representations. 
Despite the rapid development mentioned above, object representation learning via a coarse-to-fine perspective for real-time UAV tracking has not been investigated yet.
Besides, most existing tracking methods~\cite{bertinetto2016fully, cao2021hift} struggle to maintain excellent performance in dynamic environments due to limited computing resources and challenges, such as partial occlusion and aspect ratio change.
Consequently, an effective progressive representation learning framework for UAV tracking is urgently needed.

\begin{figure*}[!t]	
    \centering
    \vspace{6pt}
    \includegraphics[width=1\linewidth]{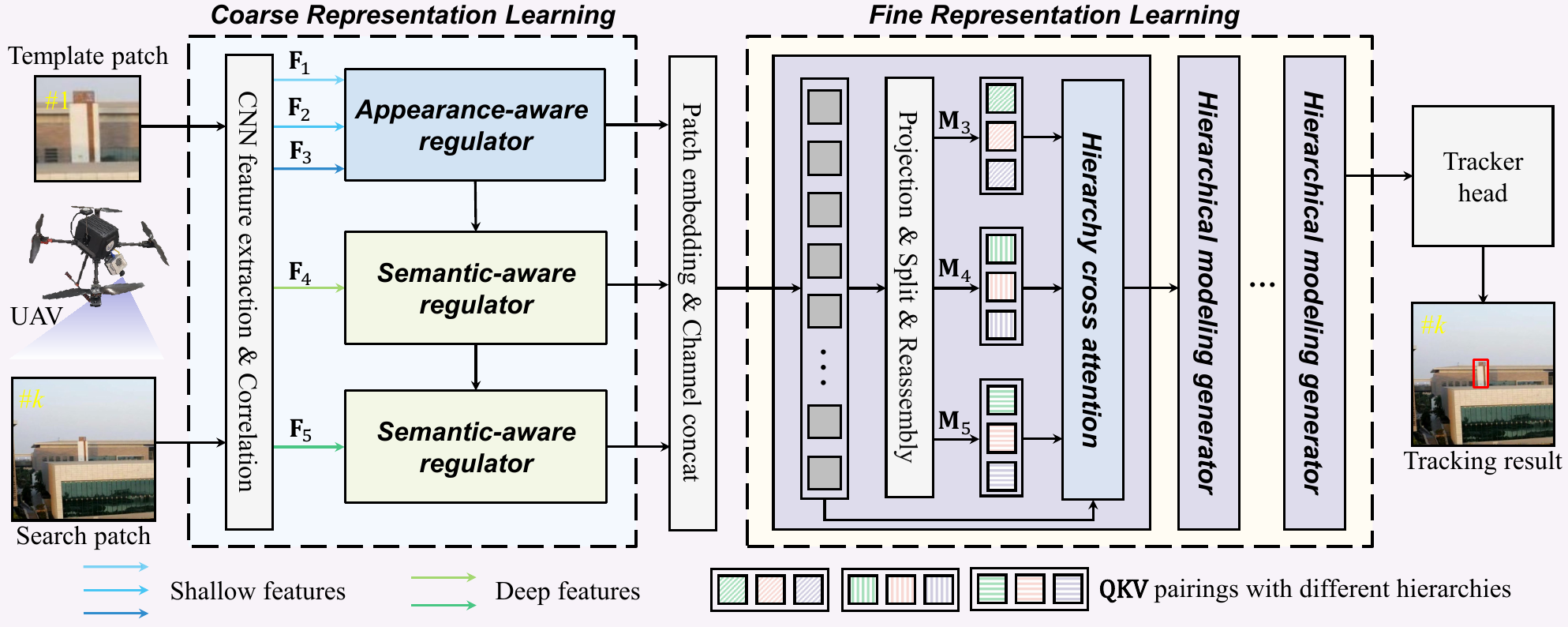}
    \vspace{-15pt}
    \caption{
    Illustration of the proposed progressive representation learning framework for UAV tracking. In the coarse representation learning, the appearance-aware regulator and semantic-aware regulator are employed to generate coarse object representations, which highlight different features of the image. In the fine representation learning, the coarse object representations are first patched, then projected, split, and reassembled to obtain $\textbf{M}_3$, $\textbf{M}_4$, and $\textbf{M}_5$ respectively, followed by fusion via hierarchical cross-attention. Best viewed in color (Image frames are from UAV123~\cite{mueller2016benchmark}).
    }  
    \label{fig:main}
    \vspace{-15pt}
\end{figure*}

\section{Proposed Method}
As depicted in Fig.~\ref{fig:main}, the proposed PRL-Track is divided into coarse representation learning and fine representation learning. 
The coarse representation learning generates coarse object representations, obtaining the local spatial information of the object.
Building upon this foundation, the fine representation learning generates robust fine object representations for UAV tracking.
With the coarse-to-fine progressive perspective, the proposed framework ensures tracking performance in complex dynamic environments, such as occlusion and aspect ratio change.

\subsection{Coarse Representation Learning}

For coarse representation learning, the CNN-based backbone is first utilized to extract multi-scale features. 
The features extracted by the shallow layers of CNNs tend to include a mass of appearance information. 
Instead, the features extracted by the deep layers of CNNs tend to enrich semantic information.
Therefore, the appearance-aware regulator and the semantic-aware regulator are proposed to process shallow features and deep features, respectively.

\subsubsection{Appearance-aware regulator (AR)}

The AR is utilized to learn appearance information such as color, edge, and shape from the shallow features. 

\begin{figure}[t]
    \centering
    \vspace{5pt}
    \includegraphics[width=1\linewidth]{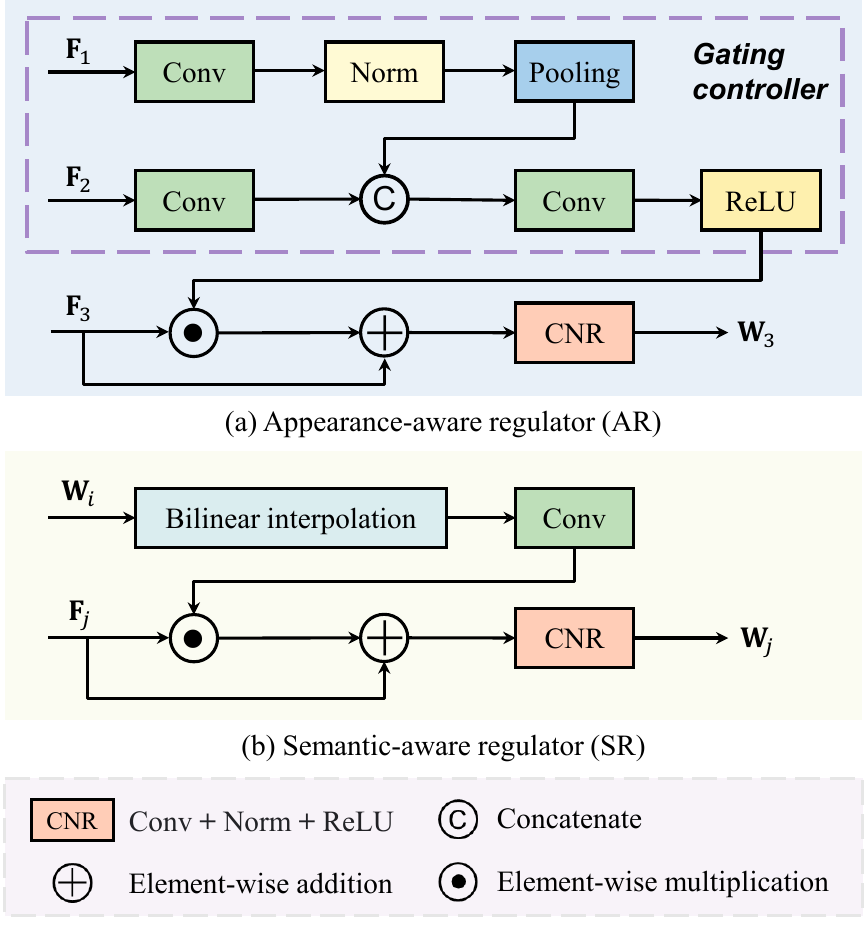}
    \vspace{-15pt}
    \caption{Structure of the proposed AR (above) and SR (below). The AR is designed to mitigate appearance inference, while the SR is designed to capture semantic information.
    }
    \label{fig:sg2}
    \vspace{-15pt}
\end{figure}

As depicted in Fig.~\ref{fig:sg2}(a), a branch called the Gating controller (GC) serves as a switch, determining the activation of related information.
Specifically, the features of the first layer $\textbf{F}_1$ and the second layer $\textbf{F}_2$ are the inputs of the GC. 
Then the convolutional operation ($\mathrm{Conv}$) is employed to achieve cross-channel information integration, where the kernel size in $\mathrm{Conv}$ is $1 \times 1$.
The intermediate results $\textbf{I}_1$ and $\textbf{I}_2$ before concatenation ($\mathrm{Concat}$) are generated as follows:
\begin{equation}
\begin{aligned}
\textbf{I}_1 &= \mathrm{Pooling}(\mathrm{Norm}(\mathrm{Conv}(\textbf{F}_1)))\ , \\
\textbf{I}_2 &= \mathrm{Conv}(\textbf{F}_2)\ ,
\end{aligned}
\end{equation}
where $\mathrm{Norm}$ denotes batch normalization, which helps stabilize and accelerate the training process. Besides, the $\mathrm{Pooling}$ operation is employed to ensure dimensional alignment.

Subsequently, a weight map $\bm{\alpha_c}$ can be obtained after $\mathrm{Concat}$ and $\mathrm{Conv}$, which is followed by a rectified linear unit activate function ($\mathrm{ReLU}$): 
\begin{equation}
\begin{aligned}
\bm{\alpha_c} = \mathrm{ReLU}{(\mathrm{Conv}(\mathrm{Concat}(\textbf{I}_1, \textbf{I}_2)))}\ .
\end{aligned}
\end{equation}

Finally, the weight map $\bm{\alpha_c}$ is employed for element-wise multiplication with the features of the third layer $\textbf{F}_3$, followed by a residual connection.
Then the output of AR, \textit{i.e.}, $\textbf{W}_3$, can be obtained as:
\begin{equation}
\begin{aligned}
    {\textbf{W}_3}  = \mathrm{CNR}(\textbf{F}_3 + \bm{\alpha_c} \cdot \textbf{F}_3)\ ,
\end{aligned}
\end{equation}
where the $\mathrm{CNR}$ represents the combination operations of $\mathrm{Conv}$, $\mathrm{Norm}$, and $\mathrm{ReLU}$. 
Additionally, residual connections and activation functions are utilized to speed up network learning and avoid the vanishing gradient problem.

\Remark The GC is employed to control the flow of features, thereby improving the quality of the object representations. 
In the learning process, $1 \times 1$ $\mathrm{Conv}$ can adaptively retain effective information or filter out redundant information, thus enhancing object representations.

\subsubsection{Semantic-aware regulator (SR)}
The $\mathrm{SR}$ is designed to learn semantic information from the deep features, \textit{i.e.}, the features from the fourth and fifth layers.

As illustrated in Fig. \ref{fig:sg2}(b), the SR takes the outputs from the previous layer $\textbf{W}_i$ and the feature of current layer $\textbf{F}_j$ as inputs.
This enables the SR to dynamically integrate contextual information from both shallow and deep features.
Then, the outputs of the two SRs used in the coarse representation learning, \textit{i.e.}, $\textbf{W}_4$ and $\textbf{W}_5$, can be obtained as:
\begin{equation}
\begin{aligned}
    {\textbf{W}_4} &= \mathrm{CNR}({\textbf{F}_4} + {\textbf{F}_4} \cdot \mathrm{Conv}(\mathrm{BLI}(\textbf{W}_3)))\ , \\
    {\textbf{W}_5} &= \mathrm{CNR}({\textbf{F}_5} + {\textbf{F}_5} \cdot \mathrm{Conv}(\mathrm{BLI} ({\textbf{W}_4})))\ ,
\end{aligned}
\end{equation}
where the $\mathrm{BLI}$ denotes bilinear interpolation, ensuring the alignment of feature dimensions. 
Notably, the first equation corresponds to the SR depicted in the upper part of Fig.~\ref{fig:main}, focusing on refining the features from the fourth layer $\textbf{F}_4$. 
Instead, the second equation corresponds to the SR depicted in the lower part of Fig.~\ref{fig:main}, which primarily enhances the features from the fifth layer $\textbf{F}_5$. 

\Remark The SR is utilized to extract useful information from deep features and transmit them to the fine representation learning. By leveraging appearance information from the AR, the SR significantly improves scene interpretation capability, which is beneficial for UAV tracking.

\subsection{Fine Representation Learning}

For fine representation learning, the hierarchical modeling generator (HMG) is designed to fuse the interaction information between coarse object representations. 
The coarse object representations generated during the previous process are first divided into patches, followed by concatenation along the channel dimension.

\begin{figure}[t]
    \vspace{6pt}
    \centering
    \includegraphics[width=1\linewidth]{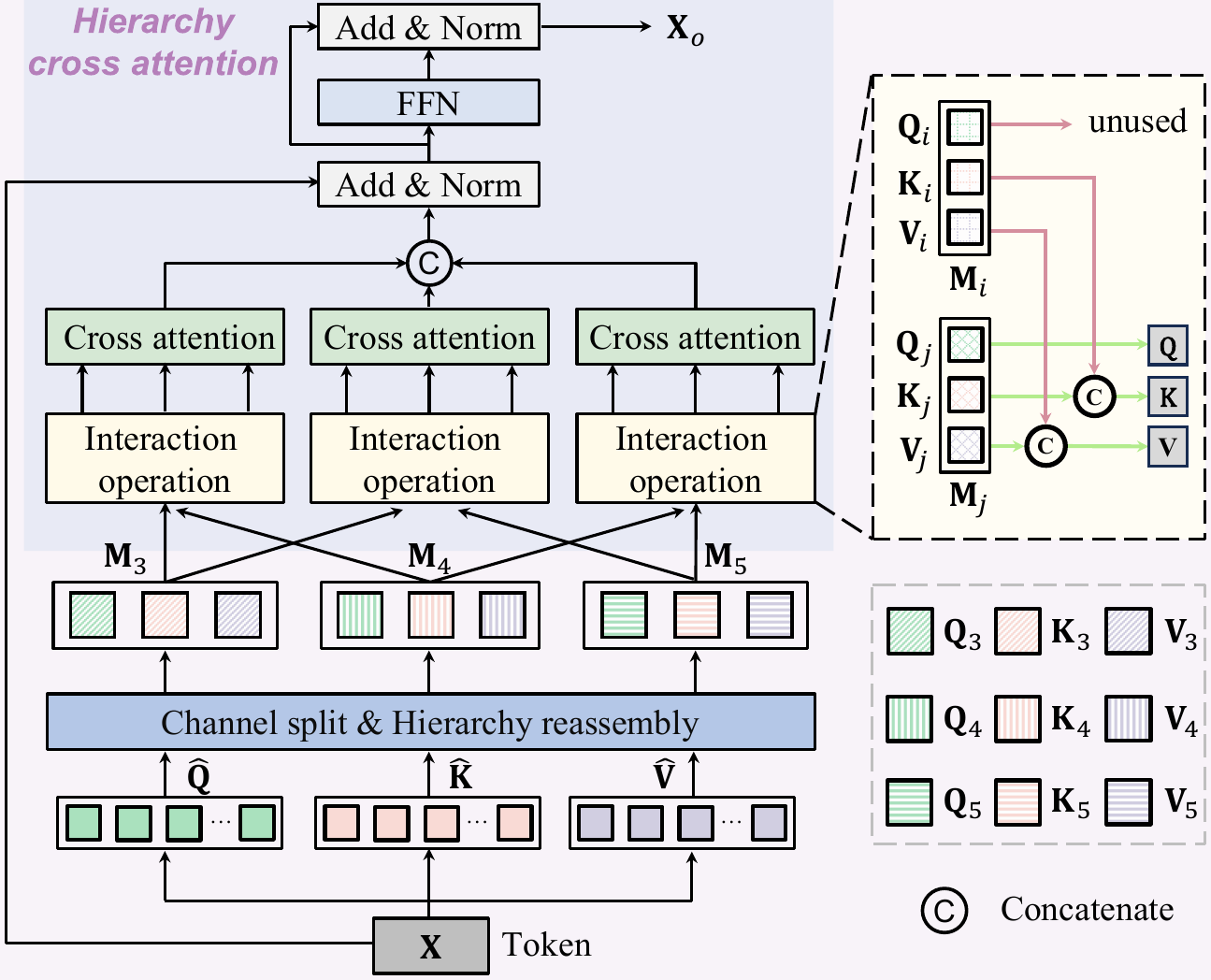}
    \vspace{-15pt}
    \caption{Detailed workflow of the proposed HMG. With interaction operation and cross-attention, the $\textbf{QKV}$ pairings with different hierarchies, \textit{i.e.}, $\textbf{M}_3$, $\textbf{M}_4$, and $\textbf{M}_5$, can communicate with each other. Best viewed in color.}
    \label{fig:sg3}
    \vspace{-15pt}
\end{figure}

As shown in Fig.~\ref{fig:sg3}, the token $\textbf{X}$ aggregated by coarse object representations is decomposed into $\textbf{QKV}$ pairings with different hierarchies, namely $\textbf{M}_3$, $\textbf{M}_4$, and $\textbf{M}_5$. 
Then they are intertwined in the ViT feature space by performing cross-attention after the interaction operation.
This strategy enables the model to capture the relationship between coarse object features at different hierarchies, thereby improving the model's representation ability.

Specifically, the process begins by decomposing the input $\textbf{X}$ into query ($\widehat{\textbf{Q}}$), key ($\widehat{\textbf{K}}$), and value ($\widehat{\textbf{V}}$) vectors via linear projection.
For the query vectors ($\widehat{\textbf{Q}}$), further splitting is conducted at the channel level, yielding $\textbf{Q}_3$, $\textbf{Q}_4$, and $\textbf{Q}_5$. Similar operations are performed for the $\widehat{\textbf{K}}$ and $\widehat{\textbf{V}}$, respectively.
From level 3 to level 5, the corresponding query, key, and value pairs at each tier are utilized to reassembly $\textbf{QKV}$ pairings, which can be represented as follows:
\begin{equation}
\begin{aligned}
\textbf{M}_i = \mathrm{Concat}(\textbf{Q}_i, \textbf{K}_i, \textbf{V}_i)\ ,\quad \mathrm{for~} i = 3, 4, 5\ .
\end{aligned}
\end{equation}

Within the proposed HMG, the hierarchy cross-attention is designed to enhance the interaction between different hierarchy representations.
To establish hierarchical connections, interaction operations are performed between $\textbf{M}_3$ and $\textbf{M}_4$, as well as between $\textbf{M}_3$ and $\textbf{M}_5$, and between $\textbf{M}_4$ and $\textbf{M}_5$.
During the interaction operation between $\textbf{M}_i$ and $\textbf{M}_j$, the keys $\textbf{K}_i$ from $\textbf{M}_i$ and $\textbf{K}_j$ from $\textbf{M}_j$ are concatenated, as well as the values $\textbf{V}_i$ and $\textbf{V}_j$, which can be expressed as:
\begin{equation}
\begin{aligned}
\textbf{K}_{ij} = \mathrm{Concat}(\textbf{K}_i, \textbf{K}_j)\ , \\
\textbf{V}_{ij} = \mathrm{Concat}(\textbf{V}_i, \textbf{V}_j)\ ,
\end{aligned}
\end{equation}
where $i < j$, $\textbf{K}_{ij}$ denotes the concatenated key from $\textbf{M}_i$ and $\textbf{M}_j$, while $\textbf{V}_{ij}$ represents the concatenated value.

Then, cross-attention mechanisms are utilized to integrate information, which can be represented as follows:
\begin{equation}
\begin{aligned}
    \textbf{H}_{att}^{34} &= \mathrm{Softmax}\left(\frac{\textbf{Q}_4 \cdot [\textbf{K}_3, \textbf{K}_4]^T}{\sqrt{\textit{d}}}\right) \cdot [\textbf{V}_3, \textbf{V}_4]\ , \\
    \textbf{H}_{att}^{35} &= \mathrm{Softmax}\left(\frac{\textbf{Q}_5 \cdot [\textbf{K}_3, \textbf{K}_5]^T}{\sqrt{\textit{d}}}\right) \cdot [\textbf{V}_3, \textbf{V}_5]\ , \\
    \textbf{H}_{att}^{45} &= \mathrm{Softmax}\left(\frac{\textbf{Q}_5 \cdot [\textbf{K}_4, \textbf{K}_5]^T}{\sqrt{\textit{d}}}\right) \cdot [\textbf{V}_4, \textbf{V}_5]\ ,
\end{aligned}
\end{equation}
where $\textit{d}$ represents the dimension of the concatenated key. Besides, $\textbf{H}_{att}^{34}$, $\textbf{H}_{att}^{35}$, and $\textbf{H}_{att}^{45}$ are the attention maps of hierarchical representations, respectively.

\Remark The fine representation learning accepts purified coarse object representations and focuses on information fusion across various hierarchical representations.
Excluding low-level queries in cross-attention streamlines the integration of relevant information across different levels of representation, thereby reducing computational costs.

Subsequently, $\textbf{H}_{att}^{34}$, $\textbf{H}_{att}^{35}$, and $\textbf{H}_{att}^{45}$ are concatenated along the channel, followed by residual connection to the input $\textbf{X}$, which can be expressed as:
\begin{equation}
\begin{aligned}
\textbf{W}_c = \mathrm{Norm}(\mathrm{Concat}(\textbf{H}_{att}^{34}, \textbf{H}_{att}^{35}, \textbf{H}_{att}^{45}) + \textbf{X})\ .
\end{aligned}
\end{equation}

Finally, the processed $\textbf{W}_c$ further undergoes adjustments through a feed-forward network ($\mathrm{FFN}$) and $\mathrm{Norm}$.
Thereby, the output of the HMG, denoted as $\textbf{X}_o$, can be expressed as:
\begin{equation}
\begin{aligned}
\textbf{X}_o = \mathrm{Norm}(\mathrm{FFN}(\textbf{W}_c) + \textbf{W}_c)\ .
\end{aligned}
\end{equation}

\Remark The strategic integration of cross-attention mechanisms facilitates precise interaction and effective fusion of diverse hierarchical features. 
Moreover, by iteratively fusing coarse object representations, the proposed HMG gradually captures both local and global information for improving performance in complex dynamic environments.

\section{Experiments}

\subsection{Implementation Details}

The proposed PRL-Track is trained using Python 3.8 and PyTorch 1.13.1 on 2 NVIDIA A100 GPUs for 70 epochs. 
The backbone of PRL-Track is initialized using AlexNet~\cite{krizhevsky2012imagenet}, which has been pre-trained on ImageNet~\cite{russakovsky2015imagenet}.
The learning rate initiates at $5 \times 10^{-4}$, rises to $10^{-2}$, and subsequently decreases to $10^{-4}$ in log space.
Additionally, the template patch is limited to dimensions of $127 \times 127 \times 3$, while the search patch is constrained to $287 \times 287 \times 3$.
The training dataset are COCO~\cite{lin2014microsoft}, GOT-10K~\cite{huang2019got}, and LaSOT~\cite{fan2019lasot}.

\begin{table*}[b]
  \centering
  \vspace{-10pt}
  \caption{Comparative evaluation of 6 SOTA trackers on UAVTrack112\_L based on attributes. The best two performances are highlighted in \textcolor{red}{red} and \textcolor{green}{green}, respectively.
  }
  \vspace{-5pt}
  \renewcommand{\arraystretch}{1.2}
  \resizebox{\linewidth}{!}{
    \colorbox{table_c}{
    \fontsize{7}{7}\selectfont
    \begin{tabular}{c|cc|cc|cc|cc}
    \toprule
    \multirow{2}{*}{\diagbox[height=3.4em]{Trackers}{Attributes}}
    &\multicolumn{2}{c|}{Aspect Ratio Change}&\multicolumn{2}{c|}{Partial Occlusion}
    &\multicolumn{2}{c|}{Scale Variation}& \multicolumn{2}{c}{Viewpoint Change} \\
    \cmidrule(lr){2-3} \cmidrule(lr){4-5} \cmidrule(lr){6-7} \cmidrule(lr){8-9}
    & 
    \multicolumn{1}{>{\centering\arraybackslash}m{1.cm}}{Prec.} & \multicolumn{1}{>{\centering\arraybackslash}m{1.cm}|}{Succ.} & \multicolumn{1}{>{\centering\arraybackslash}m{1.cm}}{Prec.} & \multicolumn{1}{>{\centering\arraybackslash}m{1.cm}|}{Succ.} & \multicolumn{1}{>{\centering\arraybackslash}m{1.cm}}{Prec.} & \multicolumn{1}{>{\centering\arraybackslash}m{1.cm}|}{Succ.} & \multicolumn{1}{>{\centering\arraybackslash}m{1.cm}}{Prec.} & \multicolumn{1}{>{\centering\arraybackslash}m{1.cm}}{Succ.} \\
    \midrule
    SE-SiamFC\cite{sosnovik2021scale} & 0.699 & 0.442 & 0.770 & 0.480 & 0.718 & 0.465 & 0.442 & 0.673 \\
    SiamAPN++~\cite{cao2021siamapn++} & 0.700 & 0.511 & 0.725 & 0.517 & 0.718 & 0.522 & 0.495 & 0.681 \\
    HiFT~\cite{cao2021hift}           & 0.712 & 0.528 & 0.760 & 0.557 & 0.721 & 0.541 & 0.491 & 0.657 \\
    SGDViT~\cite{10161487}            & 0.719 & 0.536 & 0.762 & 0.560 & 0.731 & 0.543 & \textcolor{green}{0.514} & \textcolor{green}{0.695} \\
    LPAT~\cite{fu2022local}           & \textcolor{green}{0.735} & \textcolor{green}{0.541} & \textcolor{green}{0.802} & \textcolor{green}{0.589} & \textcolor{green}{0.749} & \textcolor{green}{0.557} & 0.502 & 0.690 \\
    \midrule
    \textbf{PRL-Track (Ours)} & \textcolor{red}{\textbf{0.780}} & \textcolor{red}{\textbf{0.582}} & \textcolor{red}{\textbf{0.819}} & \textcolor{red}{\textbf{0.607}} & \textcolor{red}{\textbf{0.795}} & \textcolor{red}{\textbf{0.591}} & \textcolor{red}{\textbf{0.542}} & \textcolor{red}{\textbf{0.738}} \\
    \bottomrule
    \end{tabular}
    }
  }
  \label{tab:all}%
\end{table*}

\subsection{Evaluation Metrics}

The one-pass evaluation (OPE) metrics \cite{mueller2016benchmark} are essential for assessing tracking performance, including precision and success rate.
Specifically, the precision is measured by the Euclidean distance between the center of the predicted box and the ground truth, which is denoted as the center location error (CLE).
The precision plot is drawn by counting the percentage of frames within a certain threshold of CLE.
In the general evaluation, the threshold for tracker ranking is set to 20 pixels.
The success rate is computed through the intersection over union (IoU) of the ground truth with the predicted box.
The success plot is drawn by counting the percentage of frames whose IoU exceeds a predetermined threshold.
Meanwhile, the area under the curve (AUC) is computed to rank trackers. 

\begin{figure*}[!t]	
    \centering
    \vspace{6pt}
    {
        \colorbox{table_c}{
        \includegraphics[width=0.32\linewidth]{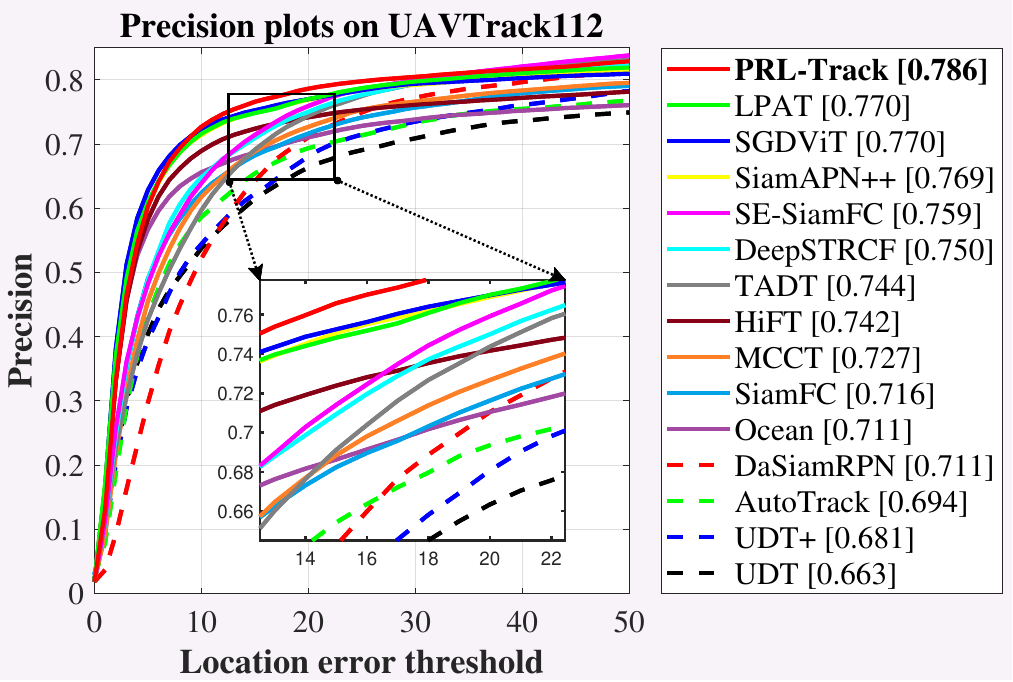}
        \includegraphics[width=0.32\linewidth]
        {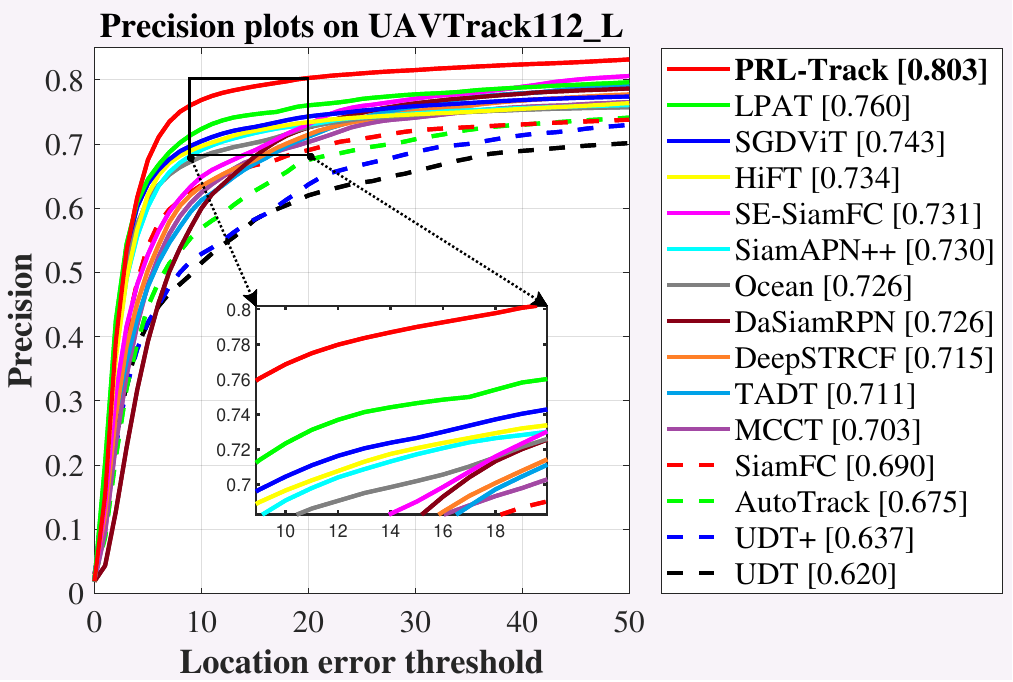}
        \includegraphics[width=0.32\linewidth]{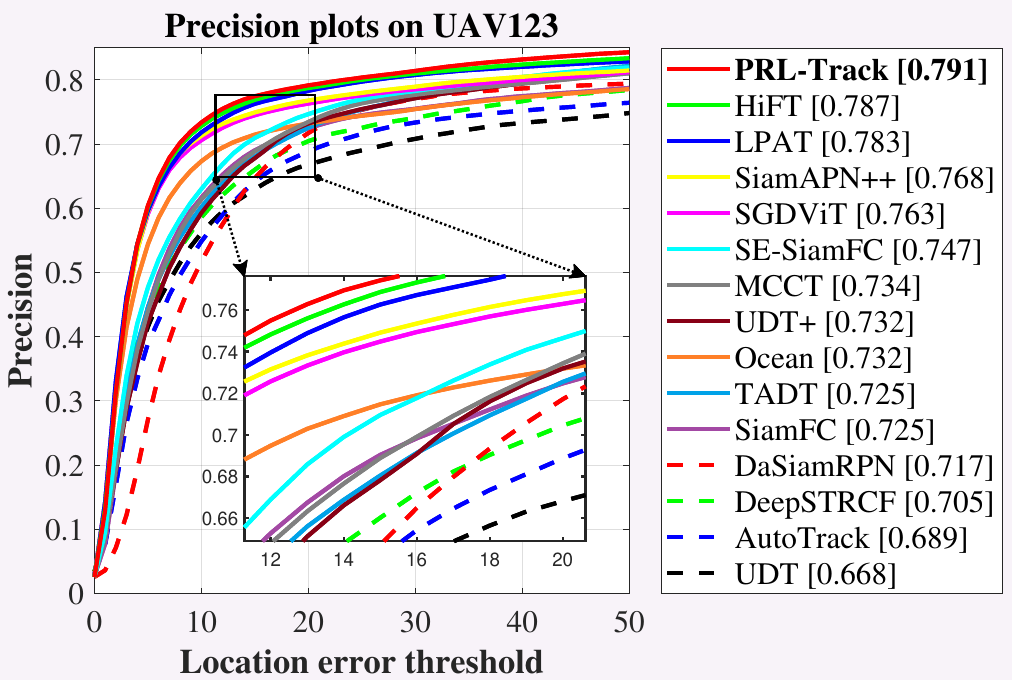}
        }
    } 
    \\[0.01em]
    \setlength{\abovecaptionskip}{-4pt}
    {
        \colorbox{table_c}
        {
        \includegraphics[width=0.32\linewidth]{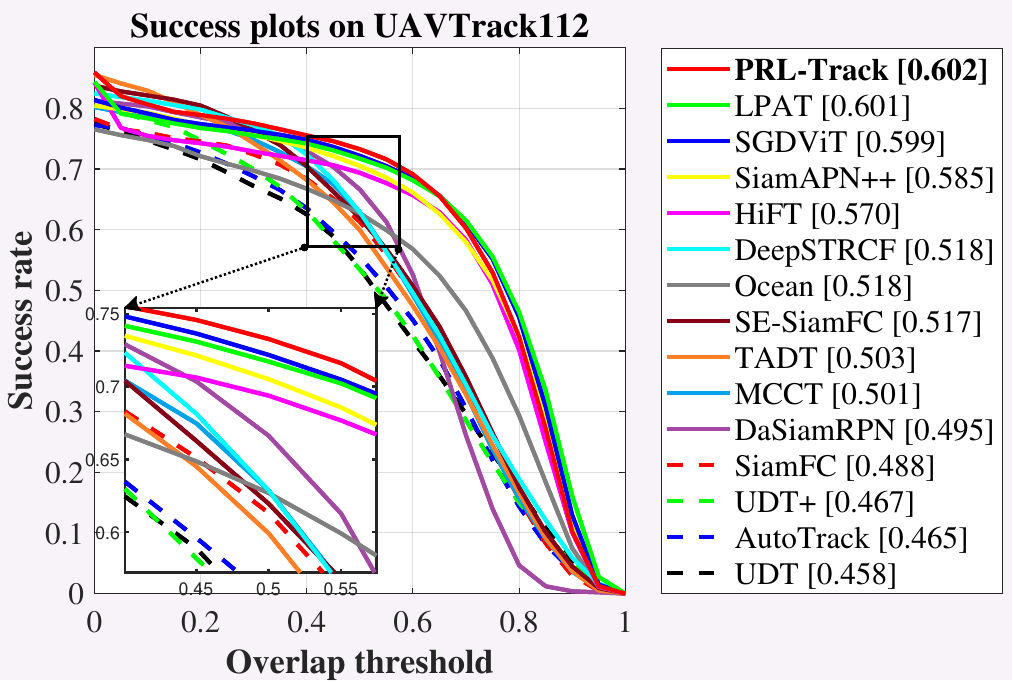}
        \includegraphics[width=0.32\linewidth]{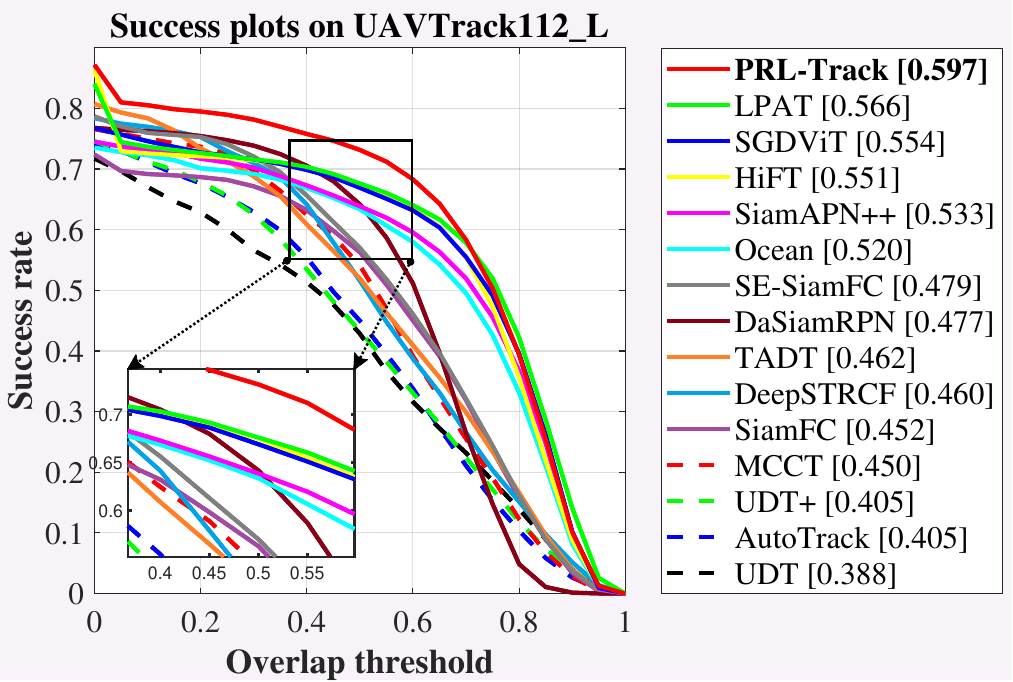}
        \includegraphics[width=0.32\linewidth]{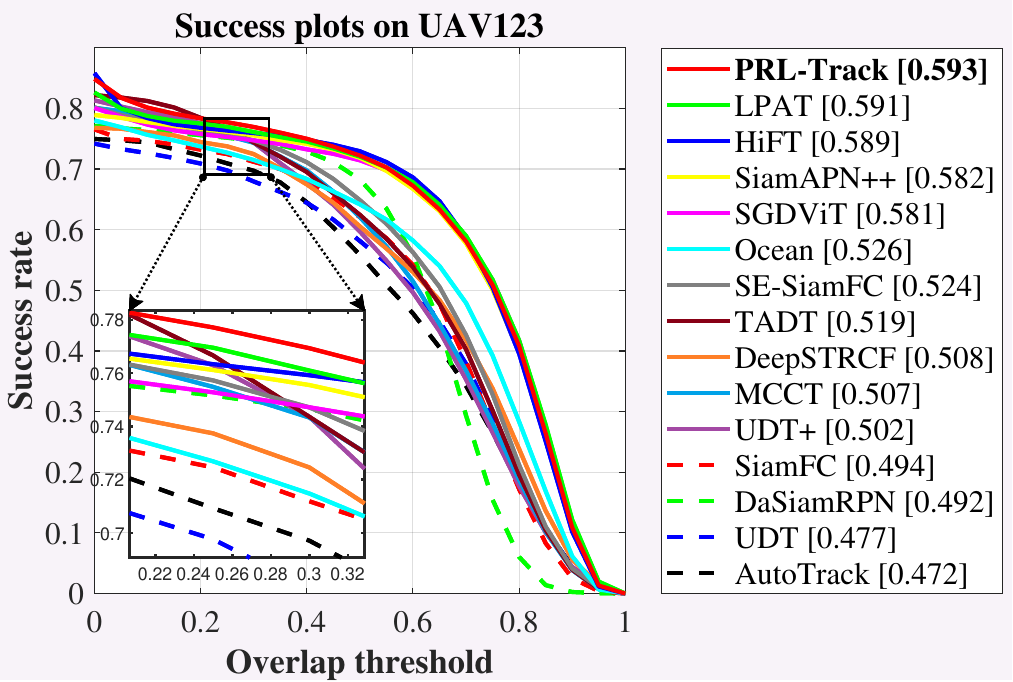}
        }
    }
    \vspace{-5pt}
    \caption
        {Overall performance of PRL-Track and SOTA trackers on UAVTrack112~\cite{fu2021onboard}, UAVTrack112\_L~\cite{fu2021onboard}, and UAV123~\cite{mueller2016benchmark}. The experimental results showcase the superior performance of the proposed PRL-Track on all benchmarks.}
	\label{fig:overall}
    \vspace{-5pt}
\end{figure*}	

\subsection{Overall Performance}
In this section, PRL-Track is tested on three challenging and authoritative UAV tracking benchmarks with other 14 existing SOTA trackers including LPAT~\cite{fu2022local}, SGDViT~\cite{10161487}, HiFT~\cite{cao2021hift}, SiamAPN++~\cite{cao2021siamapn++}, SiamFC~\cite{bertinetto2016fully}, DeepSTRCF~\cite{Li2018STRCF}, Ocean~\cite{zhang2020ocean}, DaSiamRPN~\cite{zhu2018distractor}, SE-SiamFC~\cite{sosnovik2021scale}, MCCT~\cite{wang2018multi}, AutoTrack~\cite{li2020autotrack}, TADT~\cite{li2019target}, UDT+~\cite{wang2019unsupervised}, and  UDT~\cite{wang2019unsupervised}. 
Notably, all Siamese trackers use the same lightweight backbone, \textit{i.e.}, AlexNet~\cite{krizhevsky2012imagenet}, for a fair comparison.

\subsubsection{\textbf{UAVTrack112}}
UAVTrack112~\cite{fu2021onboard} is specifically constructed for UAV tracking, encompassing 112 sequences that introduce challenges for real-world evaluations.
It encompasses common challenges~\cite{mueller2016benchmark} encountered in UAV tracking, including aspect ratio change, similar objects, partial occlusion, and so on.
The results shown in Fig.~\ref{fig:overall} demonstrate the remarkable performance of PRL-Track, attaining precision (\textbf{0.786}) and success rates (\textbf{0.602}).

\subsubsection{\textbf{UAVTrack112\_L}}
UAVTrack112\_L~\cite{fu2021onboard} consists of 45 long-term tracking sequences and includes over 60K frames in total.
Fig.~\ref{fig:overall} demonstrates that PRL-Track yields the best performance compared with other SOTA trackers.
In the precision, PRL-Track leads the pack with a remarkable score of \textbf{0.803}, surpassing LPAT (0.760) and SGDViT (0.743), which trail behind in second and third place, respectively.
Similarly, PRL-Track achieves the top success rate of \textbf{0.597}, outperforming LPAT (0.566) and SGDViT (0.554).

\Remark In this work, UAVTrack112\_L is utilized to validate the long-term tracking performance of the proposed PRL-Track. The experimental results indicate that PRL-Track performs exceptionally well on long sequences, providing a more stable and sustained tracking capability.

\subsubsection{\textbf{UAV123}}
UAV123~\cite{mueller2016benchmark} consists of 123 challenging sequences with a combined total of over 112K frames.
These sequences involve demanding aerial scenarios, encompassing occlusion, illumination variation, and low-resolution challenges.
Performance evaluation on UAV123 offers valuable insights into the advancement of aerial visual tracking.
As shown in Fig.~\ref{fig:overall}, PRL-Track stands out from other trackers with a success rate (\textbf{0.791}) and precision (\textbf{0.593}).

\subsection{Attribute-Based Comparison}

The robustness of PRL-Track in handling complex UAV tracking challenges is evaluated through attribute-based comparisons.
Specifically, the attributes of aspect ratio change (ARC), partial occlusion (POC), scale variation (SV), and viewpoint change (VC) are considered during the evaluation process. 
As illustrated in TABLE~\ref{tab:all}, PRL-Track performs the best in all four attributes compared with the other 5 SOTA trackers.
Notably, PRL-Track achieves superior performance in the ARC, surpassing the second-best performance by \textbf{4.5\%} in precision, and achieving \textbf{4.1\%} increase in success rate.
This substantial improvement demonstrates that the proposed PRL-Track can exploit the global connection of multi-scale features, thereby better adapting to scenarios where the scale of the tracking object changes.
Additionally, when confronted with partial occlusion, the ViT-based HMG utilizes purified object representations for global modeling, mitigating the impact of object feature degradation caused by occlusion.
Moreover, when encountering scale variation, the progressive process of coarse-to-fine exploration can generate more discriminative object representations to keep reliable tracking.

\Remark The promising results demonstrate that the proposed PRL-Track can learn robust object representations to tackle the challenging scenarios mentioned above. Moreover, these robust object representations contribute to the effectiveness of long-term tracking.

\subsection{Ablation Study}

To demonstrate the effectiveness of each representation learning within PRL-Track, detailed studies conducted on UAVTrack112\_L are presented in this section. 
To ensure fairness, each variant of the tracker is configured with the same settings (including training strategy and parameter configurations) except for the studied module.

\subsubsection{Clarification of symbol}
First, the symbols used in TABLE~\ref{tab:ab} are explained.
This work considers the model with only feature extraction and regression \& classification network as Baseline.
FLP represents the fine representation learning.
AR and SR represent different components used in the coarse representation learning.
PRL-Track denotes the full version of the proposed progressive representation learning framework.

\subsubsection{Result analysis}
As presented in the TABLE~\ref{tab:ab}, integrating FLP directly into the Baseline significantly improved its performance, improving precision by about 10.09\% and success rate by 13.16\%. 
This is attributed to the hierarchy modeling generator, which facilitates the integration of features across various scales.
However, combining the SR and FLP can lead to performance degradation due to appearance interference from shallow features. 
On the other hand, combining the AR and FLP enhances tracking precision by 13.11\%. 
Furthermore, adopting the Baseline+AR+SR+FLP configuration yields the best performance, showcasing improvement in precision by \textbf{15.71\%} and in success rate by \textbf{17.29\%} compared to the Baseline.
All the aforementioned results verify the efficiencies of the coarse representation learning (AR+SR) and FLP in improving object representation exploration for UAV tracking.

\subsection{Qualitative Evaluation}
\begin{figure}[t]	
    \centering
    \vspace{6pt}
    \colorbox{table_c}{
    \includegraphics[width=0.945\linewidth]{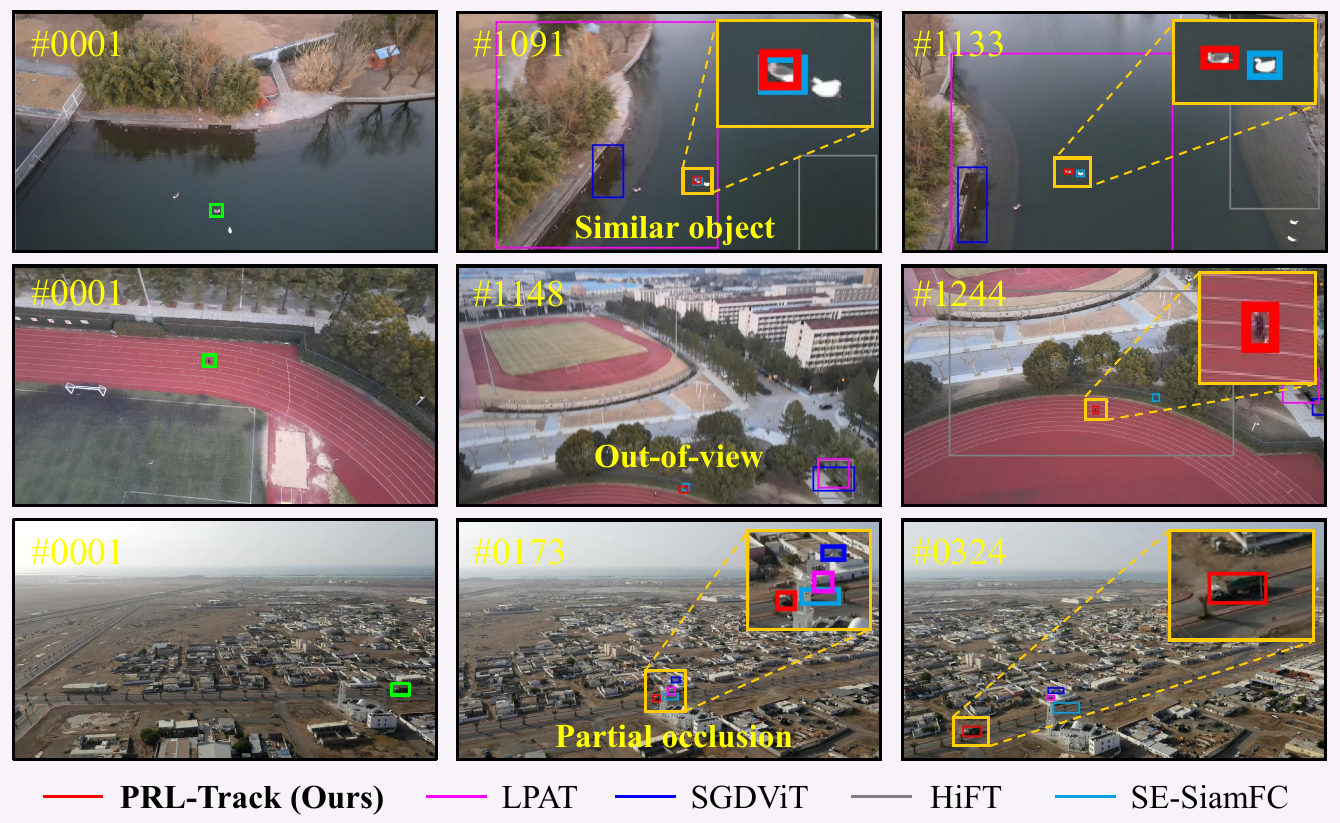}
    }
    \vspace{-15pt}
    \caption{Qualitative comparison of the proposed PRL-Track with other 4 SOTA trackers on three challenging UAV tracking sequences: duck1\_2 and jogging2 from UAVTrack112~\cite{fu2021onboard}, and truck2 from UAV123~\cite{mueller2016benchmark}. The \textcolor{green}{green} box in the first frame of each sequence represents the tracking object.
    }
    \label{fig:evaluation}
    \vspace{-15pt}
\end{figure}

As shown in Fig.~\ref{fig:evaluation}, the visualization comparison results between PRL-Track and the other 4 SOTA trackers demonstrate the robustness of PRL-Track in complex dynamic environments.
When encountering similar objects during the tracking process, the two learning processes within PRL-Track produce discriminative object representations, enabling stable and reliable tracking. 
In contrast, SE-SiamFC~\cite{sosnovik2021scale} is disrupted by similar objects, leading to tracking failure.
Furthermore, as observed from the second row of Fig.~\ref{fig:evaluation}, only PRL-Track completes the re-detection task and achieves tracking restoration after a brief out-of-view period.
Finally, in the common scenario of occlusion encountered in UAV tracking, PRL-Track also exhibits superior performance.
Owing to the robust fine object representations, the proposed PRL-Track achieves reliable tracking performance.

\section{Real-World Tests}
\begin{figure}[t]	
    \centering
    \vspace{6pt}
    \colorbox{table_c}{
    \includegraphics[width=0.94\linewidth]{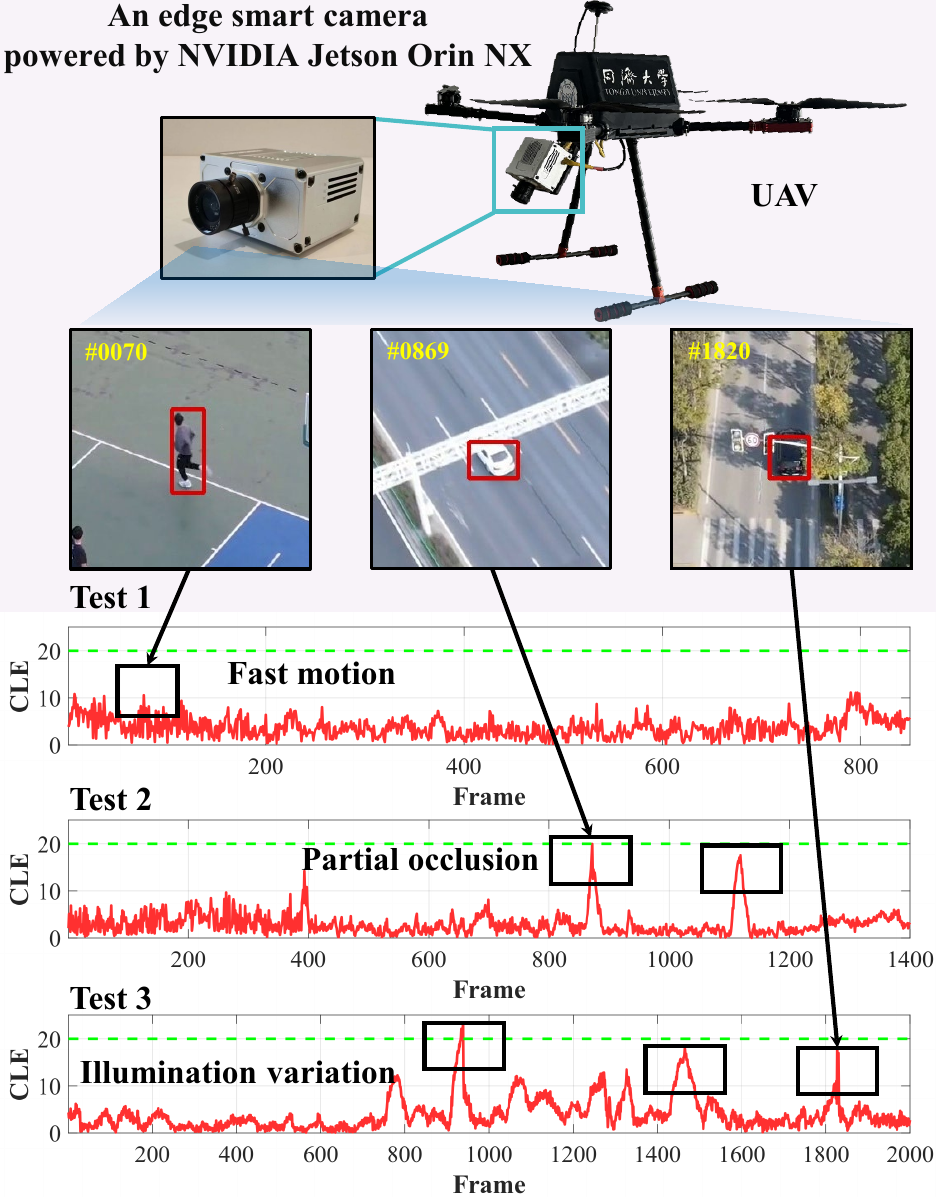}
    }
    \vspace{-15pt}
    \caption{Visualization of real-world tests: the \textcolor{red}{red} bounding boxes denote tracking results. The center location error (CLE) score below 20 is deemed reliable in the real-world test.}
    \label{fig:real}
    \vspace{-15pt}
\end{figure}

To demonstrate the real-world applicability of PRL-Track, extensive testing is conducted on a typical UAV platform, as shown in Fig.~\ref{fig:real}.
Specifically, the UAV platform is equipped with an NVIDIA Jetson Orin NX 16GB-based edge smart camera.
During the testing phase, the edge smart camera exhibits the following average utilization rates: RAM usage is at 32.67\%, while GPU and CPU record average utilizations of 28.81\% and 14.15\%, respectively.
The experimental results from several of these tests are shown in Fig.~\ref{fig:real}.
These sequences present a variety of challenges, including fast motion, partial occlusion, and illumination variation.

In Test 1, the tracked object engages in a basketball game on the court, characterized by rapid and frequent movements. 
Additionally, due to shooting actions, bodily deformation occurs intermittently. 
Nonetheless, the PRL-Track consistently maintains a high level of tracking precision in such dynamic scenarios.
The Test 2 and Test 3 sequences focus on tracking cars during steady flights, including scenarios with partial occlusion and illumination variation. 
When encountering occlusion, minor fluctuations are observed in the tracking results but quickly restore stability.
Furthermore, the Test 3 sequence highlights the performance of PRL-Track over extended durations, showcasing its robustness in long-term tracking scenarios.
Finally, the proposed PRL-Track remains a speed exceeding \textbf{42.6} frames per second, demonstrating its superior tracking speed.
The experiment results in real-world tests underscore the ability of PRL-Track to learn object representations and achieve stable tracking. 

\begin{table}[b]
  \centering
  \vspace{-10pt}
  \caption{Ablation study of the proposed framework on UAVTrack112\_L. $\Delta$ shows improvement over Baseline.
  }
  \vspace{-5pt}
  \renewcommand{\arraystretch}{1.2} 
  \resizebox{\linewidth}{!}{
    \colorbox{table_c}{
    \begin{tabular}{lcccc}
    \hline
    Trackers & Prec. & $\Delta_{Prec.}$ (\%) & Succ. & $\Delta_{Succ.}$ (\%) \\
    \hline
    Baseline & 0.694 & - & 0.509 & -  \\
    Baseline+FLP & 0.764 & +10.09 & 0.576 &  +13.16 \\
    Baseline+SR+FLP & 0.777 & +11.96 & 0.567 & +11.39 \\
    Baseline+AR+FLP & 0.785 & +13.11 & 0.577 & +13.36 \\
    \hline
    \textbf{PRL-Track}  & \textbf{0.803} & \textbf{+15.71} & \textbf{0.597} & \textbf{+17.29} \\
    \hline
    \end{tabular}
    }
    }
  \label{tab:ab}%
\end{table}
   
\section{Conclusions}

In this work, a novel progressive representation learning framework, \textit{i.e.}, PRL-Track, is proposed to extract robust object representations for UAV tracking.
In the proposed PRL-Track, two CNN-based regulators are utilized to create coarse object representations. 
Furthermore, the ViT-based hierarchical modeling generator is adopted to exploit coarse object representations.
This progressive learning process empowers the tracker, \textit{i.e.}, PRL-Track, to generate robust object representations, thereby better addressing the challenges in complex UAV scenarios.
Extensive experiments, including challenging real-world tests, demonstrate that PRL-Track has achieved outstanding performance.
We are convinced that our framework can promote further research in UAV tracking and foster related practical applications.





\section*{Acknowledgement}
This work is supported by the National Natural Science Foundation of China (No. 62173249) and the Natural Science Foundation of Shanghai (No. 20ZR1460100).

\balance

\bibliographystyle{IEEEtran}
\bibliography{ref}

\begin{thebibliography}{10}
\providecommand{\url}[1]{#1}
\csname url@samestyle\endcsname
\providecommand{\newblock}{\relax}
\providecommand{\bibinfo}[2]{#2}
\providecommand{\BIBentrySTDinterwordspacing}{\spaceskip=0pt\relax}
\providecommand{\BIBentryALTinterwordstretchfactor}{4}
\providecommand{\BIBentryALTinterwordspacing}{\spaceskip=\fontdimen2\font plus
\BIBentryALTinterwordstretchfactor\fontdimen3\font minus \fontdimen4\font\relax}
\providecommand{\BIBforeignlanguage}[2]{{%
\expandafter\ifx\csname l@#1\endcsname\relax
\typeout{** WARNING: IEEEtran.bst: No hyphenation pattern has been}%
\typeout{** loaded for the language `#1'. Using the pattern for}%
\typeout{** the default language instead.}%
\else
\language=\csname l@#1\endcsname
\fi
#2}}
\providecommand{\BIBdecl}{\relax}
\BIBdecl

\bibitem{10341486}
J.~He, Z.~Sun, N.~Cao, D.~Ming, and C.~Cai, ``{Target Attribute Perception Based UAV Real-Time Task Planning in Dynamic Environments},'' in \emph{Proceedings of the IEEE/RSJ International Conference on Intelligent Robots and Systems (IROS)}, 2023, pp. 888--895.

\bibitem{10341725}
J.~Pak, B.~Kim, C.~Ju, S.~H. You, and H.~I. Son, ``{UAV-Based Trilateration System for Localization and Tracking of Radio-Tagged Flying Insects: Development and Field Evaluation},'' in \emph{Proceedings of the IEEE/RSJ International Conference on Intelligent Robots and Systems (IROS)}, 2023, pp. 4981--4988.

\bibitem{mcarthur2020pose}
D.~R. McArthur, Z.~An, and D.~J. Cappelleri, ``{Pose-Estimate-Based Target Tracking for Human-Guided Remote Sensor Mounting with A UAV},'' in \emph{Proceedings of the IEEE International Conference on Robotics and Automation (ICRA)}, 2020, pp. 10\,636--10\,642.

\bibitem{cao2021siamapn++}
Z.~Cao, C.~Fu, J.~Ye, B.~Li, and Y.~Li, ``{SiamAPN++: Siamese Attentional Aggregation Network for Real-Time UAV Tracking},'' in \emph{Proceedings of the IEEE/RSJ International Conference on Intelligent Robots and Systems (IROS)}, 2021, pp. 3086--3092.

\bibitem{fu2023siamese}
C.~Fu, K.~Lu, G.~Zheng, J.~Ye, Z.~Cao, B.~Li, and G.~Lu, ``{Siamese Object Tracking for Unmanned Aerial Vehicle: A Review and Comprehensive Analysis},'' \emph{Artificial Intelligence Review}, vol.~56, no. Suppl 1, pp. 1417--1477, 2023.

\bibitem{wang2019unsupervised}
N.~Wang, Y.~Song, C.~Ma, W.~Zhou, W.~Liu, and H.~Li, ``{Unsupervised Deep Tracking},'' in \emph{Proceedings of the IEEE/CVF Conference on Computer Vision and Pattern Recognition (CVPR)}, 2019, pp. 1308--1317.

\bibitem{fu2024SAMDA}
C.~Fu, L.~Yao, H.~Zuo, G.~Zheng, , and J.~Pan, ``{SAM-DA: UAV Tracks Anything at Night with SAM-Powered Domain Adaptation},'' in \emph{Proceedings of the IEEE International Conference on Advanced Robotics and Mechatronics (ICARM)}, 2024, pp. 1--8.

\bibitem{krizhevsky2012imagenet}
A.~Krizhevsky, I.~Sutskever, and G.~E. Hinton, ``{ImageNet Classification with Deep Convolutional Neural Networks},'' in \emph{Proceedings of the Advances in Neural Information Processing Systems (NIPS)}, vol.~25, 2012, pp. 1097--1105.

\bibitem{he2016deep}
K.~He, X.~Zhang, S.~Ren, and J.~Sun, ``{Deep Residual Learning for Image Recognition},'' in \emph{Proceedings of the IEEE Conference on Computer Vision and Pattern Recognition (CVPR)}, 2016, pp. 770--778.

\bibitem{9857212}
J.~Fang, H.~Lin, X.~Chen, and K.~Zeng, ``{A Hybrid Network of CNN and Transformer for Lightweight Image Super-Resolution},'' in \emph{Proceedings of the IEEE/CVF Conference on Computer Vision and Pattern Recognition Workshops (CVPRW)}, 2022, pp. 1102--1111.

\bibitem{dosovitskiy2020image}
A.~Dosovitskiy, L.~Beyer, A.~Kolesnikov, D.~Weissenborn, X.~Zhai, T.~Unterthiner, M.~Dehghani, M.~Minderer, G.~Heigold, S.~Gelly \emph{et~al.}, ``{An Image is Worth 16x16 Words: Transformers for Image Recognition at Scale},'' in \emph{Proceedings of the International Conference on Learning Representations (ICLR)}, 2020, pp. 1--22.

\bibitem{cao2021hift}
Z.~Cao, C.~Fu, J.~Ye, B.~Li, and Y.~Li, ``{HiFT: Hierarchical Feature Transformer for Aerial Tracking},'' in \emph{Proceedings of the IEEE/CVF International Conference on Computer Vision (ICCV)}, 2021, pp. 15\,457--15\,466.

\bibitem{10040235}
Z.~Peng, Z.~Guo, W.~Huang, Y.~Wang, L.~Xie, J.~Jiao, Q.~Tian, and Q.~Ye, ``{Conformer: Local Features Coupling Global Representations for Recognition and Detection},'' \emph{IEEE Transactions on Pattern Analysis and Machine Intelligence}, vol.~45, no.~8, pp. 9454--9468, 2023.

\bibitem{8546079}
L.~Zhang, Y.~Dong, and Y.~Wu, ``{Multi-Layer CNN Features Aggregation for Real-Time Visual Tracking},'' in \emph{Proceedings of the International Conference on Pattern Recognition (ICPR)}, 2018, pp. 2404--2409.

\bibitem{8850096}
M.~Zhao, S.~Zhong, X.~Fu, B.~Tang, and M.~Pecht, ``{Deep Residual Shrinkage Networks for Fault Diagnosis},'' \emph{IEEE Transactions on Industrial Informatics}, vol.~16, no.~7, pp. 4681--4690, 2020.

\bibitem{10143709}
K.~Li, Y.~Wang, J.~Zhang, P.~Gao, G.~Song, Y.~Liu, H.~Li, and Y.~Qiao, ``{UniFormer: Unifying Convolution and Self-Attention for Visual Recognition},'' \emph{{IEEE Transactions on Pattern Analysis and Machine Intelligence}}, vol.~45, no.~10, pp. 12\,581--12\,600, 2023.

\bibitem{raghu2021vision}
M.~Raghu, T.~Unterthiner, S.~Kornblith, C.~Zhang, and A.~Dosovitskiy, ``{Do Vision Transformers See Like Convolutional Neural Networks?}'' in \emph{Proceedings of the Advances in Neural Information Processing Systems (NIPS)}, vol.~34, 2021, pp. 12\,116--12\,128.

\bibitem{bertinetto2016fully}
L.~Bertinetto, J.~Valmadre, J.~F. Henriques, A.~Vedaldi, and P.~H. Torr, ``{Fully-Convolutional Siamese Networks for Object Tracking},'' in \emph{Proceedings of the European Conference on Computer Vision Workshops (ECCVW)}, 2016, pp. 850--865.

\bibitem{guo2017learning}
Q.~Guo, W.~Feng, C.~Zhou, R.~Huang, L.~Wan, and S.~Wang, ``{Learning Dynamic Siamese Network for Visual Object Tracking},'' in \emph{Proceedings of the IEEE International Conference on Computer Vision (ICCV)}, 2017, pp. 1763--1771.

\bibitem{li2020autotrack}
Y.~Li, C.~Fu, F.~Ding, Z.~Huang, and G.~Lu, ``{AutoTrack: Towards High-Performance Visual Tracking for UAV with Automatic Spatio-Temporal Regularization},'' in \emph{Proceedings of the IEEE/CVF Conference on Computer Vision and Pattern Recognition (CVPR)}, 2020, pp. 11\,923--11\,932.

\bibitem{wang2018multi}
N.~Wang, W.~Zhou, Q.~Tian, R.~Hong, M.~Wang, and H.~Li, ``{Multi-Cue Correlation Filters for Robust Visual Tracking},'' in \emph{Proceedings of the IEEE Conference on Computer Vision and Pattern Recognition (CVPR)}, 2018, pp. 4844--4853.

\bibitem{li2018high}
B.~Li, J.~Yan, W.~Wu, Z.~Zhu, and X.~Hu, ``{High Performance Visual Tracking with Siamese Region Proposal Network},'' in \emph{Proceedings of the IEEE Conference on Computer Vision and Pattern Recognition (CVPR)}, 2018, pp. 8971--8980.

\bibitem{chen2021transformer}
X.~Chen, B.~Yan, J.~Zhu, D.~Wang, X.~Yang, and H.~Lu, ``{Transformer Tracking},'' in \emph{Proceedings of the IEEE/CVF Conference on Computer Vision and Pattern Recognition (CVPR)}, 2021, pp. 8126--8135.

\bibitem{10161487}
L.~Yao, C.~Fu, S.~Li, G.~Zheng, and J.~Ye, ``{SGDViT: Saliency-Guided Dynamic Vision Transformer for UAV Tracking},'' in \emph{Proceedings of the IEEE International Conference on Robotics and Automation (ICRA)}, 2023, pp. 3353--3359.

\bibitem{6472238}
Y.~Bengio, A.~Courville, and P.~Vincent, ``{Representation Learning: A Review and New Perspectives},'' \emph{IEEE Transactions on Pattern Analysis and Machine Intelligence}, vol.~35, no.~8, pp. 1798--1828, 2013.

\bibitem{zou2023object}
Z.~Zou, K.~Chen, Z.~Shi, Y.~Guo, and J.~Ye, ``{Object Detection in 20 Years: A Survey},'' \emph{Proceedings of the IEEE}, vol. 111, no.~3, pp. 257--276, 2023.

\bibitem{wang2020deep}
J.~Wang, K.~Sun, T.~Cheng, B.~Jiang, C.~Deng, Y.~Zhao, D.~Liu, Y.~Mu, M.~Tan, X.~Wang \emph{et~al.}, ``{Deep High-Resolution Representation Learning for Visual Recognition},'' \emph{IEEE Transactions on Pattern Analysis and Machine Intelligence}, no.~10, pp. 3349--3364, 2020.

\bibitem{li2021esvit}
C.~Li, J.~Yang, P.~Zhang, M.~Gao, B.~Xiao, X.~Dai, L.~Yuan, and J.~Gao, ``{Efficient Self-Supervised Vision Transformers for Representation Learning},'' in \emph{Proceedings of the International Conference on Learning Representations (ICLR)}, 2022, pp. 1--27.

\bibitem{cai2023marlin}
Z.~Cai, S.~Ghosh, K.~Stefanov, A.~Dhall, J.~Cai, H.~Rezatofighi, R.~Haffari, and M.~Hayat, ``{MARLIN: Masked Autoencoder for Facial Video Representation Learning},'' in \emph{Proceedings of the IEEE/CVF Conference on Computer Vision and Pattern Recognition (CVPR)}, 2023, pp. 1493--1504.

\bibitem{mueller2016benchmark}
M.~Mueller, N.~Smith, and B.~Ghanem, ``{A Benchmark and Simulator for UAV Tracking},'' in \emph{Proceedings of the European Conference on Computer Vision (ECCV)}, 2016, pp. 445--461.

\bibitem{russakovsky2015imagenet}
O.~Russakovsky, J.~Deng, H.~Su, J.~Krause, S.~Satheesh, S.~Ma, Z.~Huang, A.~Karpathy, A.~Khosla, M.~Bernstein \emph{et~al.}, ``{ImageNet Large Scale Visual Recognition Challenge},'' \emph{International Journal of Computer Vision}, vol. 115, pp. 211--252, 2015.

\bibitem{lin2014microsoft}
T.-Y. Lin, M.~Maire, S.~Belongie, J.~Hays, P.~Perona, D.~Ramanan, P.~Doll{\'a}r, and C.~L. Zitnick, ``{Microsoft COCO: Common Objects in Context},'' in \emph{Proceedings of the European Conference on Computer Vision (ECCV)}, 2014, pp. 740--755.

\bibitem{huang2019got}
L.~Huang, X.~Zhao, and K.~Huang, ``{GOT-10K: A Large High-Diversity Benchmark for Generic Object Tracking in the Wild},'' \emph{IEEE Transactions on Pattern Analysis and Machine Intelligence}, vol.~43, no.~5, pp. 1562--1577, 2019.

\bibitem{fan2019lasot}
H.~Fan, L.~Lin, F.~Yang, P.~Chu, G.~Deng, S.~Yu, H.~Bai, Y.~Xu, C.~Liao, and H.~Ling, ``{LaSOT: A High-Quality Benchmark for Large-Scale Single Object tracking},'' in \emph{Proceedings of the IEEE/CVF Conference on Computer Vision and Pattern Recognition (CVPR)}, 2019, pp. 5374--5383.

\bibitem{sosnovik2021scale}
I.~Sosnovik, A.~Moskalev, and A.~W. Smeulders, ``{Scale Equivariance Improves Siamese Tracking},'' in \emph{Proceedings of the IEEE/CVF Winter Conference on Applications of Computer Vision (WACV)}, 2021, pp. 2765--2774.

\bibitem{fu2022local}
C.~Fu, W.~Peng, S.~Li, J.~Ye, and Z.~Cao, ``{Local Perception-Aware Transformer for Aerial Tracking},'' in \emph{Proceedings of the IEEE/RSJ International Conference on Intelligent Robots and Systems (IROS)}, 2022, pp. 12\,122--12\,129.

\bibitem{fu2021onboard}
C.~Fu, Z.~Cao, Y.~Li, J.~Ye, and C.~Feng, ``{Onboard Real-Time Aerial Tracking with Efficient Siamese Anchor Proposal Network},'' \emph{IEEE Transactions on Geoscience and Remote Sensing}, pp. 1--13, 2021.

\bibitem{Li2018STRCF}
F.~Li, C.~Tian, W.~Zuo, L.~Zhang, and M.~H. Yang, ``{Learning Spatial-Temporal Regularized Correlation Filters for Visual Tracking},'' in \emph{Proceedings of the IEEE/CVF Conference on Computer Vision and Pattern Recognition (CVPR)}, 2018, pp. 4904--4913.

\bibitem{zhang2020ocean}
Z.~Zhang, H.~Peng, J.~Fu, B.~Li, and W.~Hu, ``{Ocean: Object-Aware Anchor-Free Tracking},'' in \emph{Proceedings of the European Conference on Computer Vision (ECCV)}, 2020, pp. 771--787.

\bibitem{zhu2018distractor}
Z.~Zhu, Q.~Wang, B.~Li, W.~Wu, J.~Yan, and W.~Hu, ``{Distractor-Aware Siamese Networks for Visual Object Tracking},'' in \emph{Proceedings of the European Conference on Computer Vision (ECCV)}, 2018, pp. 101--117.

\bibitem{li2019target}
X.~Li, C.~Ma, B.~Wu, Z.~He, and M.-H. Yang, ``{Target-Aware Deep Tracking},'' in \emph{Proceedings of the IEEE/CVF Conference on Computer Vision and Pattern Recognition (CVPR)}, 2019, pp. 1369--1378.

\end{thebibliography}

\end{document}